\documentclass[runningheads]{llncs}

\usepackage{graphicx}
\usepackage{amsmath}
\usepackage{hyperref}
\usepackage{ragged2e}
\usepackage{booktabs} 
\usepackage[T1]{fontenc}
\usepackage{graphicx}
\begin{document}
\title{Optimized Custom CNN for Real-Time Tomato Leaf Disease Detection}
%
%
\author{Mangsura Kabir Oni\inst{1} \and
Tabia Tanzin Prama\inst{2}}
%

%
\institute{Jahangirnagar University, Dhaka-1342, Bangladesh. 
\email{tabiatanzin1234@gmail.com}\\
} 

\maketitle              
\begin{abstract}
In Bangladesh, tomatoes are a staple vegetable, prized for their versatility in various culinary applications. However, the cultivation of tomatoes is often hindered by a range of diseases that can significantly reduce crop yields and quality. Early detection of these diseases is crucial for implementing timely interventions and ensuring the sustainability of tomato production. Traditional manual inspection methods, while effective, are labor-intensive and prone to human error. To address these challenges, this research paper sought to develop an automated disease detection system using Convolutional Neural Networks (CNNs). A comprehensive dataset of tomato leaves was collected from the Brahmanbaria district, preprocessed to enhance image quality, and then applied to various deep learning models. Comparative performance analysis was conducted between YOLOv5, MobileNetV2, ResNet18, and our custom CNN model. In our study, the Custom CNN model achieved an impressive accuracy of 95.2\%, significantly outperforming the other models, which achieved an accuracy of 77\%, 89.38\% and 71.88\% respectively. While other models showed solid performance, our Custom CNN demonstrated superior results specifically tailored for the task of tomato leaf disease detection. These findings highlight the strong potential of deep learning techniques for improving early disease detection in tomato crops. By leveraging these advanced technologies, farmers can gain valuable insights to detect diseases at an early stage, allowing for more effective management practices. This approach not only promises to boost tomato yields but also contributes to the sustainability and resilience of the agricultural sector, helping to mitigate the impact of plant diseases on crop production.

\end{abstract}
\section{Introduction}
Agriculture is essential for feeding the world, and new technologies have constantly made farming better. Artificial Intelligence (AI) is one of the most important advancements, which helps farmers grow more crops, use resources wisely, and make farming easier and more efficient. AI-driven solutions are widely used in agriculture for tasks like optimizing irrigation systems, predicting crop yields, and automating pest control. These technologies rely on machine learning (ML) algorithms, which can process vast amounts of agricultural data, allowing farmers to make better decisions in real time and use resources more efficiently.One of the most promising applications of AI in agriculture is computer vision, which has revolutionized plant health monitoring. By using deep learning (DL), a specialized field of AI, computer vision techniques allow for automated, highly accurate plant disease detection. Traditional disease detection methods depend on human experts visually inspecting plants, which can be slow, expensive, and sometimes inaccurate due to human error. In contrast, AI-powered computer vision analyzes images of plant leaves to detect signs of disease quickly and accurately. This method not only saves time but also makes disease detection more accessible to farmers who may not have specialized training in plant pathology. For instance, we can mention some papers: Aashua, Kanchan Rajwarb, Millie Panta Kusum Deepa(2024) show a comprehensive review of how machine learning (ML) techniques are applied to optimize crop output and minimize environmental impact. The paper discusses the role of ML in analyzing agricultural data to improve farm productivity and profitability [13]. Another paper titled “Artificial Intelligence in Sustainable Vertical Farming" (2023) explores the integration of AI in vertical farming systems. The paper highlights how AI applications, including machine learning, computer vision, and robotics, optimize resource usage, automate tasks, and enhance decision-making in controlled environment agriculture [14].
Tomato (Solanum lycopersicum) is one of the most widely cultivated crops in the world, and its productivity heavily depends on the health of its leaves. Leaves play a vital role in photosynthesis, respiration, and transpiration, which are essential processes for plant growth and fruit production. However, tomato plants are vulnerable to various challenges, including pests, nutrient deficiencies, and environmental stressors such as extreme temperatures and drought. Among these challenges, foliar diseases, diseases that affect the leaves pose a serious threat to crop yield and quality. Common tomato leaf diseases include mosaic virus, yellow leaf curl virus, target spot, two-spotted spider mite, septoria leaf spot, leaf mold, late blight, early blight, and bacterial spot. These diseases can spread quickly, damaging the plant and reducing the quality and quantity of the harvest.Early detection of these diseases is crucial for effective management. If diseases are identified in their early stages, farmers can take immediate action using strategies such as targeted fungicide applications, crop rotation, and planting disease-resistant varieties. However, identifying diseases through traditional visual inspection is not always practical, especially for large-scale farms. It requires expert knowledge, takes a lot of time, and is prone to mistakes. This is where deep learning (DL) comes in as a game-changing solution for plant disease detection.Deep learning, a subset of machine learning, uses multi-layered neural networks to recognize patterns in complex datasets, such as images. It has been widely adopted in various fields, including healthcare, finance, and autonomous vehicles, and is now proving to be highly effective in agriculture. One of the most commonly used deep learning techniques for image analysis is Convolutional Neural Networks (CNNs). CNNs are particularly well-suited for detecting plant diseases from leaf images because they can automatically learn features such as color changes, spots, and texture variations associated with disease symptoms.This research focuses on using CNN-based models to detect tomato leaf diseases accurately and efficiently. The study aims to build and analyze a system that can distinguish between healthy and diseased leaves using advanced deep-learning techniques. To achieve this goal, a comprehensive dataset of tomato leaf images will be created, capturing leaves under different conditions, including various lighting, angles, and backgrounds, to ensure a robust and diverse training dataset. The study will also compare different CNN models, including a Custom CNN, MobileNetV2, ResNet18, and YOLO v5, to determine which model provides the best performance. Each model will be fine-tuned by optimizing hyperparameters to improve accuracy and efficiency. The models will be evaluated using standard performance metrics such as accuracy, precision, recall, and F1 score to ensure their reliability in real-world applications. By comparing different models, this research will provide insights into the most effective approach for detecting tomato leaf diseases.The ultimate goal of this study is to develop an AI-powered tool that farmers can use to quickly and accurately detect tomato leaf diseases. By integrating deep learning into agriculture, farmers will be able to take early action against plant diseases, reducing crop losses and improving yield. This technology will not only save time and resources but also promote sustainable farming by minimizing the need for excessive pesticide use. By making disease detection more accessible and efficient, this research aims to support farmers in growing healthier crops, increasing food production, and ensuring global food security.

\section{Literature Review}
Several researches on detecting and classifying tomato leaf diseases were done. Numerous studies have been carried out in order to forecast and classify leaf diseases in various tomato trees. This research focused on binary classification of tomato leaves whether they are healthy or diseased and used a range of deep learning methods to deal with them. This chapter shows a summary of the pertinent work being efficiently completed by a number of professionals in the relevant sector. The suggested CNN model exhibited an accuracy of 91.2\%, which was noticeably higher than pre-trained CNN models like VGG16 (77.2\%), Mobilenet (63.75\%), and Inception (63.4\%), according to a research by Agarwal et al. (2020)[1]. The study focused on using convolutional neural networks to identify diseases in tomato leaves. For this investigation, they suggested using a CNN model with three convolution layers and three max pooling layers. The advantages of not employing a pre-trained model are further highlighted by the study's discovery that the recommended model required far less storage space—just 1.5 MB—than the pre-trained models, which needed 100 MB. In 2020, Hatuwal, Shakya, and Joshi [2] conducted research on the detection of plant diseases using various machine learning models, including Support Vector Machine (SVM), K-nearest Neighbour (KNN), Random Forest Classifier (RFC), and Convolution Neural Network (CNN). The CNN model had the highest accuracy of any machine learning model, scoring 97.89\%; the RFC model came in second, scoring 87.436\%; SVM came in third, scoring 78.61\%; and KNN, scoring 76.969\%. In contrast to earlier research, this work used f1-score, precision, and recall to assess its models. However, accuracy was the only factor considered when selecting the top-performing model in the final model comparison. For the purpose of classifying tomato leaf diseases using field datasets, Rajasree Rajamohanan and Beulah Christalin Latha [3] exhibited YOLOv5 in 2023. According to the test dataset, they achieved a noteworthy accuracy rate of 93\%. The early tomato leaf spot detection approach, which was created using the MobileNetv2-YOLOv3 techniques, achieves greater accuracy and real-time tomato leaf spot detection stability. Model detection efficacy was assessed using the F1 score and AP value, and testing was conducted in comparison to SSD and Faster RCNN techniques. As demonstrated by the experiment findings, the suggested model has a much better detection effect [4]. In a study on plant disease identification, Madhulatha and Ramadevi (2020) employed a deep Convolutional Neural Network model, and the suggested work was demonstrated to yield an accuracy of 96.50\%. The study uses the well-known AlexNet architecture to categorize the various plant diseases. Known for being utilized in the majority of picture classification use case scenarios, the AlexNet architecture is a Neural Network with eight layers of learnable features. The plant village dataset, which includes 54,323 photos of plant diseases across 38 distinct disease categories, is the source of the dataset used in this study [5]. The suggested work yielded 11 with an overall accuracy of 76.59\%, according to study on the identification of illnesses in paddy leaves with the KNN classifier done by Suresha, Shreekanth, and Thirumalesh (2017) on a database of 330 photos of paddy leaves. The study's sole parameter for assessing the KNN classifier was accuracy; other metrics, such precision, recall, and f1-score, which will be the subject of this work, were not used [6]. An improved Faster RCNN was used by Zhang et al. (2020) to identify four different disease categories and healthy tomato leaves. For image feature extraction, they used a depth residual network rather than VGG16, and for bounding box clustering, they applied the k-means clustering algorithm. Having a detection time of just 470 ms, experimental results on public datasets demonstrated an average identification accuracy of 98.54\% [7].

\section{Proposed Methodology}

\begin{figure}
\centering
\includegraphics[width=\textwidth]{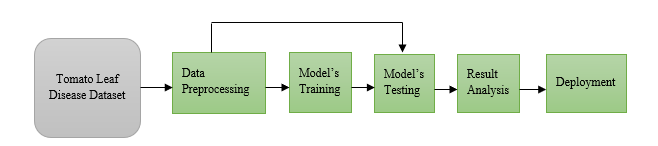}
\caption{Workflow of our proposed mobile application for tomato leaf disease identification} \label{fig1}
\end{figure}

The proposed approach follows a series of systematic steps. Figure\ref{fig1} illustrates a detailed representation of the implementation procedure. A binary classification dataset of healthy and diseased tomato leaves was preprocessed (resizing, JPEG conversion, and data augmentation) to boost training robustness. A custom CNN and transfer learning models (MobilenetV2.Yolov5 and ResNet18 with ImageNet weights) were trained on both original and augmented data. Grad-CAM visualizations, an ablation study, and callbacks for monitoring validation loss were used to optimize and select the best model. Finally, the top-performing model was deployed in a web-based system and an Android app for effective disease prediction.

\subsection{Dataset Description}
\begin{figure}
\centering
\includegraphics[width=0.7\textwidth]{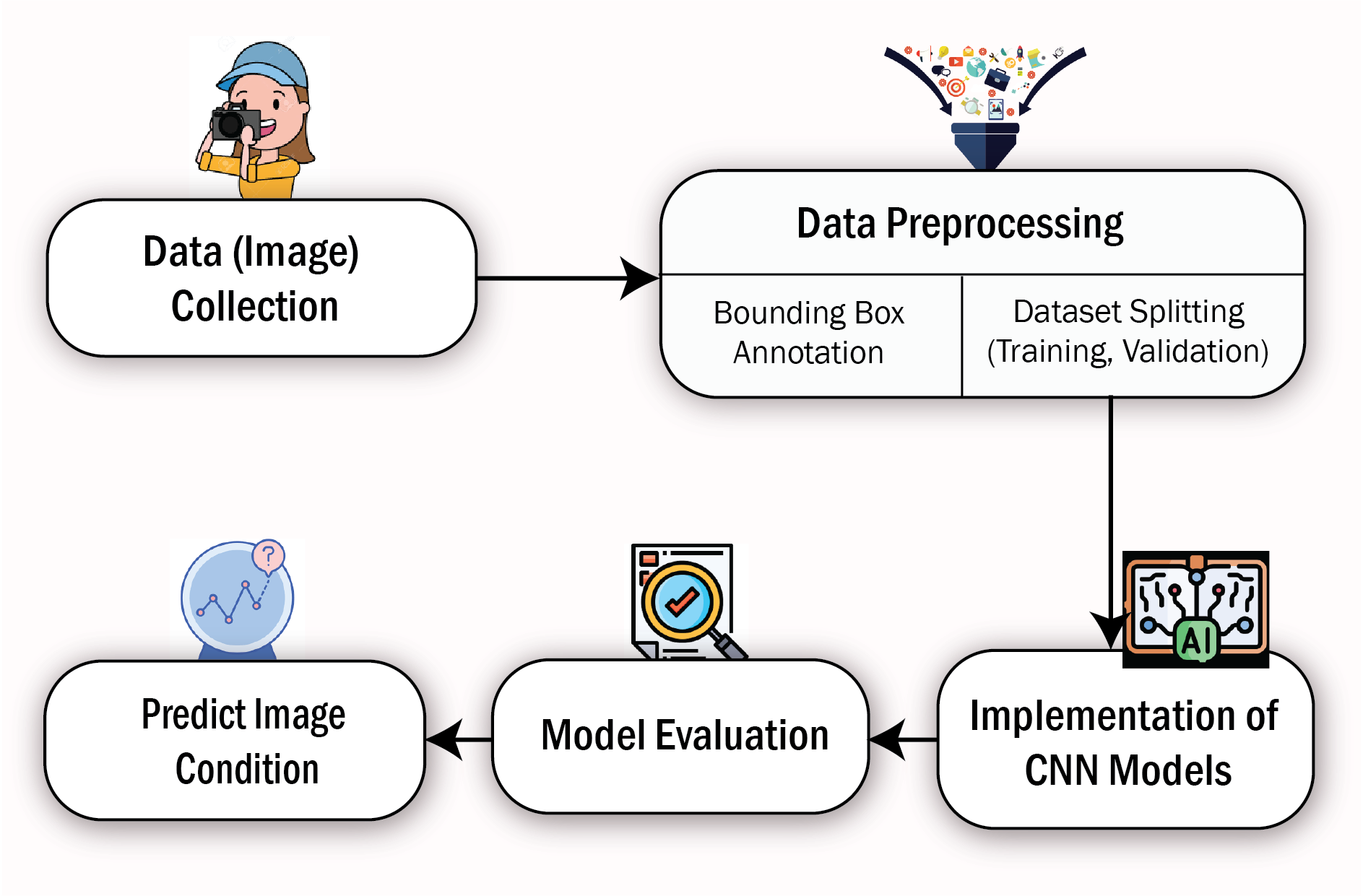}
\caption{Workflow of proposed methodology} \label{fig2}
\end{figure}

\begin{figure}
\centering
\includegraphics[width=0.7\textwidth]{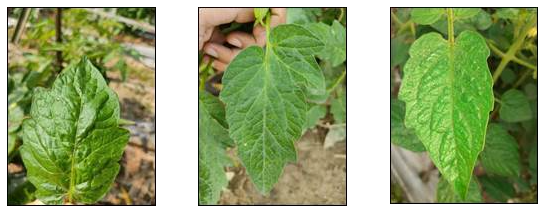}
\caption{Example of collected healthy leaf images from field} \label{fig3}
\end{figure}

\begin{figure}
\centering
\includegraphics[width=0.7\textwidth]{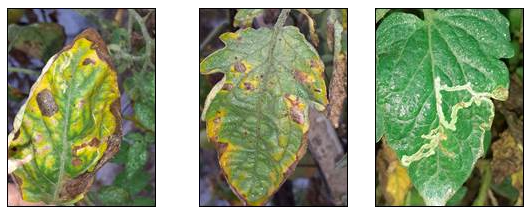}
\caption{Example of collected diseased leaf images from field} \label{fig4}
\end{figure}

We collected tomato leaf images from fields, which were located in Mohammadpur, Brahmanbaria, Bangladesh. Images of tomato plants were photographed and collected using two smartphones: Redmi Note 10 Pro and Samsung Galaxy A10 [8]. Here, Figure \ref{fig2} shows the workflow of proposed methodology  for tomato disease detection. Healthy leaves, leaves with various tomato leaf diseases, and leaves affected by common environmental stresses were captured with the consent of garden owners. The images were gathered in early February, 2024. The outdoor environment was sunny one day and foggy one day according to the daytime. On the sunny day, the temperature was 26-29 degrees Celsius. And on the foggy day, it was 17-18 degrees Celsius. We’ve collected the images of 482 healthy leaves and 546 diseased leaves. And shifted them in two folders named ‘Healthy’ and ‘Diseased’ respectively [8]. From these two folders some samples are attached below through Figure \ref{fig3} and Figure \ref{fig4}

\subsection{Data Preprocessing}
\begin{figure}
\centering
\includegraphics[width=0.7\textwidth]{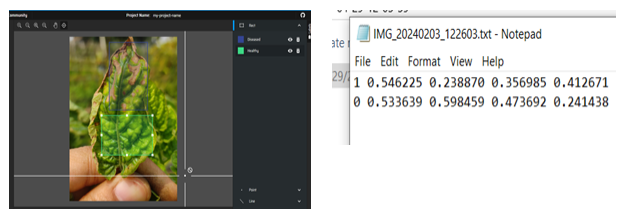}
\caption{Example of data annotating and labeling  the raw images} \label{fig5}
\end{figure}

Preparing is an essential first step to prepare our image data before feeding it into machine learning models. Before picture data is ready to be put into a computer vision model, it must be cleaned. Preprocessing images can help speed up model inference and reduce training time. There are both technical and performance reasons why preprocessing is essential. For our research project the preprocessing included the following considerations:
\begin{itemize}
    \item \textbf{Bounding Box Annotation:} We annotated our images with bounding boxes for each leaf we wanted the model to detect. The YOLO annotation format (containing class labels and bounding box coordinates) was used for labeling. We used the annotating software \href{https://www.makesense.ai/}{MakeSense} to label the images manually. By creating bounding boxes, we labeled the images, and after exporting, we obtained the labels in a text file, as illustrated in Figure \ref{fig5}. 
\end{itemize}

\begin{itemize}
    \item \textbf{Data Splitting:} After labeling our dataset, next task is splitting. Among them, approximately 80\% data was used for training, 15\% data for validation and 5\% for testing. Here 80\% indicates 810 images , 15\% indicates 160 images and 5\% indicates 60 images.

\end{itemize}

\section{Background Study}
\subsection{Convolutional Neural Networks (CNNs)}
CNNs are specialized neural networks designed for analyzing visual data. They automatically learn hierarchical features such as edges and patterns through their layers. Key components of CNNs include convolution layers, which use filters to create feature maps, and pooling layers, which reduce data dimensions while preserving critical information. Max pooling is common for extracting prominent features. Activation functions like ReLU introduce non-linearity, allowing the network to learn complex patterns. Flattened feature maps are sent to fully connected layers for classification, with SoftMax or Sigmoid functions generating probabilities for each class. Techniques such as dropout prevent overfitting, and batch normalization speeds up training by normalizing inputs. Custom CNN architectures achieve high accuracy in image classification, object detection, and segmentation. for 13 seconds
Convolutional Neural Networks (CNNs) are specialized neural architectures designed to analyze visual data by automatically learning hierarchical features from raw images. Their structure typically includes convolutional layers that apply filters to extract features (e.g., edges and textures), pooling layers (like max pooling) that reduce spatial dimensions, and fully connected layers that integrate these features for classification. Techniques such as ReLU activation, dropout, and batch normalization further enhance the model's ability to learn complex patterns and generalize well. These elements collectively enable CNNs to excel in tasks like image classification, object detection, and segmentation [12].

\subsection{YOLOv5}
YOLOv5, introduced in 2020, is an advanced object detection model ideal for tomato leaf disease detection. It incorporates dynamic anchor boxes and spatial pyramid pooling to better detect small objects. Its architecture includes a CSPDarknet backbone for feature extraction, a PANet neck enhanced with Spatial Pyramid Pooling and BottleNeckCSP for feature aggregation, and a YOLO head for predicting bounding boxes, scores, and classes. The model uses SiLU and Sigmoid activations and optimizes training with Binary Cross Entropy and CIoU loss [9].

\subsection{MobileNetV2}
MobileNetV2 is a lightweight CNN architecture designed for mobile and embedded vision applications. It balances model size and accuracy through features like depthwise separable convolutions, inverted residuals, linear bottlenecks, and squeeze-and-excitation (SE) blocks, which reduce computational complexity while preserving performance. These enhancements enable efficient processing while capturing detailed visual features, making MobileNetV2 ideal for resource-constrained devices. It is widely used in image classification, object detection, and real-time computer vision, offering high accuracy and low computational cost [11]. 

\subsection{ResNet18}
ResNet-18 is a deep residual network designed to improve gradient flow and mitigate the vanishing gradient problem through residual learning. Its 18 layers, built around residual blocks with convolutional layers, Batch Normalization, and ReLU activations, utilize shortcut connections to enhance feature extraction and training efficiency. Despite its compact size, ResNet-18 performs robustly in tasks like image classification and object detection [10].

\subsection{Evaluation Metrics}
\justify
Evaluation metrics---accuracy, precision, recall, and F1-score---are essential for assessing model performance [12]. Accuracy is defined as:
\begin{equation}
    \text{Accuracy} = \frac{TP + TN}{TP + TN + FP + FN}
\end{equation}
which measures the proportion of correct predictions. Precision,
\begin{equation}
    \text{Precision} = \frac{TP}{TP + FP}
\end{equation}
indicates the correctness of positive predictions, while recall (or True Positive Rate),
\begin{equation}
    \text{Recall} = \frac{TP}{TP + FN}
\end{equation}
reflects the model's ability to capture actual positives. The F1-score, defined as the harmonic mean of precision and recall, is given by:
\begin{equation}
    \text{F1-Score} = 2 \times \frac{\text{Precision} \times \text{Recall}}{\text{Precision} + \text{Recall}}
\end{equation}
This metric ensures a balanced evaluation, especially in scenarios with class imbalance.

\section{System Architecture of Proposed Model}

\begin{figure}
\centering
\includegraphics[width=\textwidth]{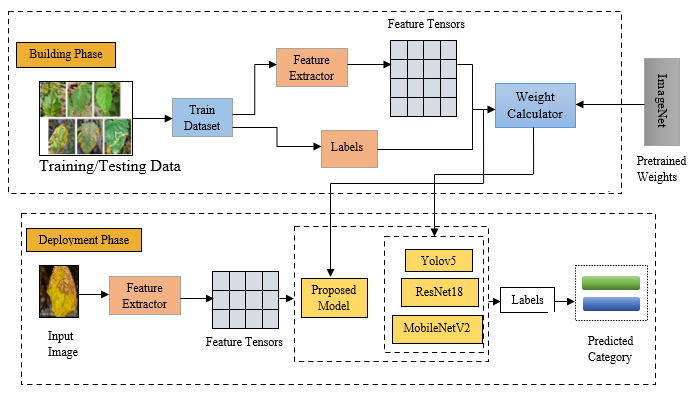}
\caption{A systematic representation of our proposed approach } \label{fig6}
\end{figure}

Figure \ref{fig6} outlines our system architecture in two phases: building and deployment. In the building phase, tomato leaf images are collected and augmented (rescaling, rotation, shifting, zooming, and flipping) before training a custom CNN for binary classification (healthy vs. diseased) using the Adam optimizer and binary cross-entropy loss. Grad-CAM is applied to highlight key regions influencing predictions. In the deployment phase, models are evaluated using accuracy, loss, and confusion matrix metrics, and the best-performing model is deployed via an interactive interface for predicting tomato leaf diseases.

\justify 

\textbf{Proposed Custom CNN:}  
The proposed custom CNN model processes preprocessed images of tomato leaves, each resized to $224 \times 224$ dimensions with three color channels, as input. The model architecture is meticulously designed to achieve high classification accuracy while maintaining computational efficiency. It comprises 4 convolutional layers, each employing a kernel size of $2 \times 2$ with \texttt{"same"} padding. These layers extract critical features from the input images. Each convolutional layer is followed by a max pooling layer, also with a kernel size of $2 \times 2$, to down-sample the feature maps and reduce spatial dimensions, effectively improving computational efficiency while retaining the most important features. Additionally, the model integrates 3 dropout layers, each with a dropout rate of 0.2 (20\%), to randomly deactivate 20\% of the neurons during training. This helps mitigate overfitting by ensuring that the model generalizes well to unseen data. The network also includes one fully connected (FC) layer, culminating in a dense output layer that employs the SoftMax activation function for multiclass classification. The weights of the custom CNN model are initialized using the Xavier Glorot uniform method, ensuring that the initial weights are drawn from a uniform distribution within the range of $\pm r$, where:
\begin{equation}
    r = \sqrt{\frac{6}{X_i + X_o}}
\end{equation}
Here, $X_i$ represents the number of input connections, and $X_o$ represents the number of output connections. This initialization prevents issues such as vanishing or exploding gradients and ensures smooth gradient flow during backpropagation.
To introduce non-linearity and accelerate learning, the Rectified Linear Unit (ReLU) activation function is employed in all hidden layers. The ReLU activation function is defined as:
\begin{equation}
    y = \max(0, x)
\end{equation}
outputting either 0 or the input value $x$. Its simplicity and computational efficiency make it a popular choice for deep learning architectures, as it avoids saturation and speeds up the convergence process.

The Adam optimizer is utilized to minimize prediction error and find optimal node weights. Adam, derived from adaptive moment estimation, combines the advantages of gradient descent with momentum and RMSProp algorithms. This optimizer adjusts the learning rate for each network weight independently, resulting in faster training, reduced memory usage, and minimal need for hyperparameter tuning. The learning rate ($\alpha$) for Adam is set at 0.001, which was determined to provide the best balance between convergence speed and model accuracy during training. The loss function used in this study is Categorical Cross-Entropy, which effectively measures the difference between true labels and predicted probabilities. Categorical Cross-Entropy combines the SoftMax activation function with the Cross-Entropy loss calculation, ensuring that the probabilities for all classes sum to 1. It penalizes predictions that deviate significantly from the true labels, encouraging the model to improve its predictive accuracy. The training process involves 100 epochs with a batch size of 32, meaning the model processes 32 images at a time before updating the weights. This batch size was chosen after an ablation study, balancing training time and model performance. The callbacks function was employed to monitor validation loss, ensuring that the best-performing model was saved as the final version. Incorporating all these factors, training parameters are represented in Table~\ref{tab:training_params}. This approach allows the custom CNN model to effectively learn from the data, achieve robust performance, and reliably classify tomato leaf diseases.

\begin{table}[h]
    \centering
    \caption{Training Parameters for the Proposed Custom CNN Model}
    \label{tab:training_params}
    \begin{tabular}{ll}
        \toprule
        \textbf{Parameter} & \textbf{Description} \\
        \midrule
        Optimization algorithm & Adam optimizer \\
        Learning rate ($\alpha$) & 0.001 \\
        Weight initialization & Xavier Glorot uniform \\
        Batch size & 32 \\
        Number of epochs & 100 \\
        Dropout rate & 0.2 (20\%) \\
        Loss function & Categorical Cross-Entropy \\
        Activation function (hidden layers) & ReLU \\
        Activation function (output layer) & SoftMax \\
        Kernel size (convolution layers) & $2 \times 2$ with "same" padding \\
        Kernel size (max pooling layers) & $2 \times 2$ \\
        \bottomrule
    \end{tabular}
\end{table}

\section{Experimental Results}
\begin{table}[h]
    \centering
    \renewcommand{\arraystretch}{1.2} 
    \setlength{\tabcolsep}{8pt} 
    \begin{tabular}{|l|c|c|c|c|}
        \hline
        \textbf{Model Name} & \textbf{Recall} & \textbf{Precision} & \textbf{F1 Score} & \textbf{Accuracy} \\ 
        \hline
        Custom CNN & 0.92 & 0.85 & 0.89 & 95.2\% \\ 
        YOLOv5 & 0.80 & 0.86 & 0.75 & 77\% \\ 
        MobileNetV2 & 0.87 & 0.90 & 0.88 & 89.38\% \\ 
        ResNet18 & 0.72 & 0.71 & 0.70 & 71.88\% \\ 
        \hline
    \end{tabular}
    \caption{Comparison among different models}
    \label{tab:comparison_models}
\end{table}

Table \ref{tab:comparison_models} shows the experimental results of models. The Custom CNN model achieved the highest performance across all metrics. It demonstrated a recall of 0.92, indicating its strong ability to correctly identify diseased leaves, and a precision of 0.85, reflecting a low rate of false positives. The model's F1 score of 0.89 showed a good balance between recall and precision, and it achieved an impressive accuracy of 95.2\%, making it the top-performing model in this task. The YOLOv5 model, while generally effective for object detection, showed a recall of 0.80, precision of 0.86, and F1 score of 0.75, with an accuracy of 77\%. These results suggest that while YOLOv5 performed reasonably well, it faced challenges in achieving high accuracy and balanced performance on this specific task. MobileNetV2 performed similarly well with a recall of 0.87, precision of 0.90, and F1 score of 0.88, achieving an accuracy of 89.38\%. This model struck a good balance between performance and efficiency, offering strong results for real-time applications in agriculture. The ResNet18 model showed the lowest performance, with a recall of 0.72, precision of 0.71, and F1 score of 0.70, resulting in an accuracy of 71.875\%. These results suggest that ResNet18 was less suitable for the task compared to the other models. Overall, the Custom CNN emerged as the best-performing model, followed by MobileNetV2, YOLOv5, and ResNet18 in terms of accuracy, recall, and F1 score. The model demonstrates exceptional performance in classifying "Healthy" and "Diseased" leaves, with an impressive overall accuracy of 95.2\%. 

\begin{figure}[h]
\centering
\includegraphics[width=0.5\textwidth]{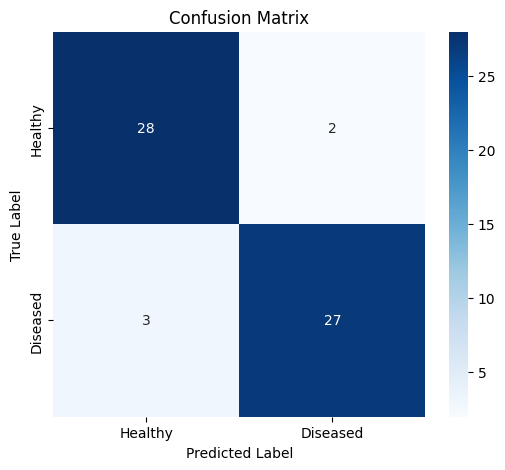}
\caption{Confusion Matrix of our proposed Custom CNN model.}
\label{fig7}
\end{figure}

The confusion matrix in Figure \ref{fig7} provides detailed insight into the classification outcomes, showing that the model correctly identified 28 out of 30 healthy instances (true positives) while misclassifying only 2 as diseased (false negatives). Similarly, it accurately predicted 27 out of 30 diseased instances (true positives) and misclassified 3 as healthy (false negatives). These results highlight the model's ability to minimize errors, particularly its strong identification of true positives in both classes.

\begin{figure}
\centering
\includegraphics[width=0.8 \textwidth]{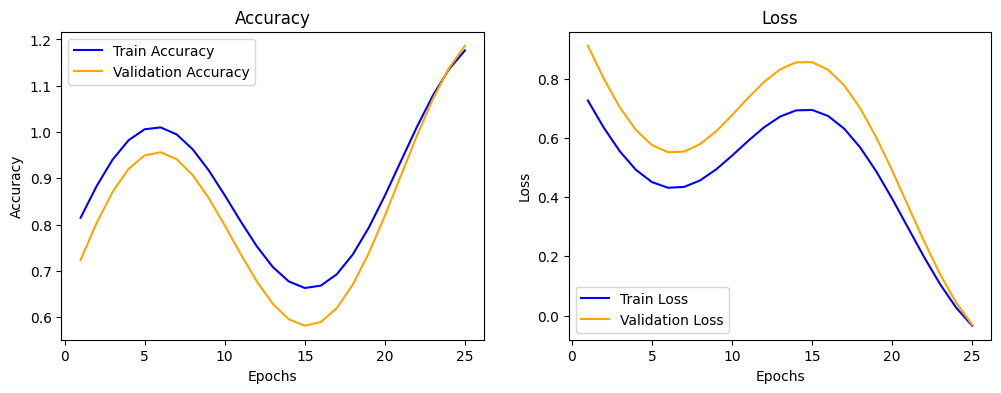}
\caption{ Accuracy and Loss of our proposefd custom CNN model} \label{fig9}
\end{figure}

The accuracy and loss graphs in Figure \ref{fig9} provide additional evidence of the model’s robustness and reliability. The training and validation accuracy curves exhibit consistent growth throughout the 25 epochs, eventually converging near the end, with the validation accuracy closely matching the training accuracy. This convergence indicates a well-trained model that avoids overfitting, as the performance on unseen validation data mirrors the training data. Meanwhile, the loss graphs demonstrate effective optimization, with the training loss steadily decreasing as the model learns, and the validation loss following a similar trend, with a slight increase mid-training before ultimately converging. This pattern is typical of a model that generalizes well without overfitting or underfitting. Overall, these results collectively highlight the model’s ability to classify healthy and diseased leaves accurately, supported by strong precision, recall, and F1 scores, as well as stable training dynamics, making it a reliable solution for binary classification tasks in leaf health analysis.


\subsection{Results and Discussion}
\begin{figure}
\centering
\includegraphics[width=0.7\textwidth]{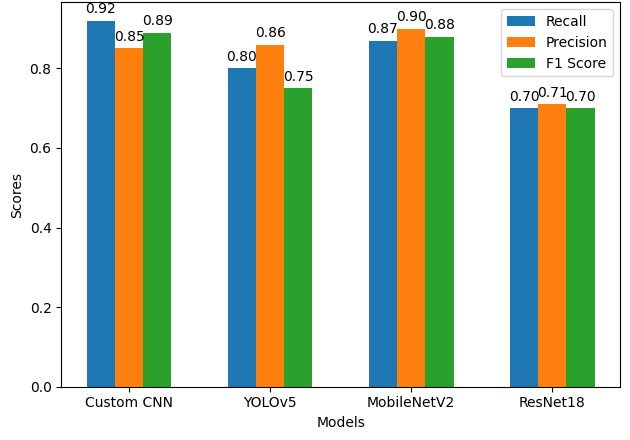}
\caption{ Comparison of different models (Custom CNN, YOLOv5, Mobilenet, Resnet18)} \label{fig11}
\end{figure}

Figure \ref{fig11} shows the performance of our proposed custom CNN model and baseline models like YOLOv5, MobileNetV2, and ResNet18 models. The custom CNN model achieved the highest accuracy of 95.2\%, along with strong recall (0.92), precision (0.85), and an F1-score of 0.89, making it well-suited for tomato leaf disease classification. YOLOv5, known for its efficiency in object detection, attained an accuracy of 77\% with a precision of 0.86. However, its recall (0.80) and F1-score (0.75) indicate that it struggled with capturing all diseased instances, which can be a limitation in real-time agricultural applications. MobileNetV2, a lightweight deep learning model, performed well with an accuracy of 89.38\%, recall of 0.87, precision of 0.90, and an F1-score of 0.88, demonstrating its effectiveness in the classification task. ResNet18, while widely used for image classification tasks, had the lowest accuracy of 71.875\%, with recall, precision, and F1-score all around 0.70, suggesting its lower suitability for tomato leaf disease detection. Overall, while YOLOv5 and MobileNetV2 offer robust performance, our custom CNN demonstrates superior accuracy and a well-balanced trade-off between recall and precision, making it the most effective model for tomato leaf disease classification in agricultural settings. 

\section{Conclusion}
This study used deep learning to classify healthy and diseased tomato leaves based on labeled images. We compared four models—YOLOv5, MobileNetV2, ResNet18, and our proposed Custom CNN model. The Custom CNN achieved the best accuracy, while YOLOv5 and MobileNetV2 also performed well. ResNet18 had weaker performance, especially in recall and precision. Limitations include the lack of data augmentation and environmental factors that affected accuracy, particularly in real-time scenarios. Future work should focus on creating a mobile app for farmers, augmenting the dataset, and improving model performance, especially in complex backgrounds, to enhance disease detection and management for healthier crops and higher yields.

\begin{credits}
\subsubsection{\ackname} We would like to express our appreciation to the farmers and agricultural experts from the Brahmanbaria district for their cooperation in providing real-world insights and facilitating the collection of tomato leaf samples. 

\end{credits}
%
%
%
%

\end{document}